\documentclass[sigconf ]{acmart}
\usepackage[ruled,vlined,linesnumbered]{algorithm2e}
\usepackage{subcaption}
\usepackage{enumitem}
\AtBeginDocument{%
  \providecommand\BibTeX{{%
    \normalfont B\kern-0.5em{\scshape i\kern-0.25em b}\kern-0.8em\TeX}}}

\copyrightyear{2024}
\acmYear{2024}
\setcopyright{acmlicensed}\acmConference[KDD '24]{Proceedings of the 30th ACM SIGKDD Conference on Knowledge Discovery and Data Mining}{August 25--29, 2024}{Barcelona, Spain}
\acmBooktitle{Proceedings of the 30th ACM SIGKDD Conference on Knowledge Discovery and Data Mining (KDD '24), August 25--29, 2024, Barcelona, Spain}
\acmDOI{10.1145/3637528.3671738}
\acmISBN{979-8-4007-0490-1/24/08}

\usepackage{amsthm}
\theoremstyle{plain}
\newtheorem{thm}{Theorem}[section] % reset theorem numbering for each chapter
\newtheorem{lem}[thm]{Lemma}
\newtheorem{co}[thm]{Corollary}

\theoremstyle{definition}
\newtheorem{defn}[thm]{Definition} % definition numbers are dependent on theorem numbers
\begin{document}

\title{PolygonGNN: Representation Learning for Polygonal Geometries with Heterogeneous Visibility Graph}

\author{Dazhou Yu}
\email{dyu62@emory.edu}
\orcid{0000-0003-2082-0834}
\affiliation{%
  \institution{Emory University}
  \city{Atlanta}
  \state{GA}
  \country{United States}
}

\author{Yuntong Hu}
\orcid{0000-0003-3802-9039}
\email{yuntong.hu@emory.edu}
\affiliation{%
  \institution{Emory University}
  \city{Atlanta}
  \state{GA}
  \country{United States}
}

\author{Yun Li}
\email{yli230@emory.edu}
\orcid{0000-0002-3205-8464}
\affiliation{%
  \institution{Emory University}
  \city{Atlanta}
  \state{GA}
  \country{United States}
}

\author{Liang Zhao}
\email{liang.zhao@emory.edu}
\orcid{0000-0002-2648-9989}
\affiliation{%
  \institution{Emory University}
  \city{Atlanta}
  \state{GA}
  \country{United States}
}


\begin{abstract}
Polygon representation learning is essential for diverse applications, encompassing tasks such as shape coding, building pattern classification, and geographic question answering. While recent years have seen considerable advancements in this field, much of the focus has been on single polygons, overlooking the intricate inner- and inter-polygonal relationships inherent in multipolygons.
To address this gap, our study introduces a comprehensive framework specifically designed for learning representations of polygonal geometries, particularly multipolygons.  Central to our approach is the incorporation of a heterogeneous visibility graph, which seamlessly integrates both inner- and inter-polygonal relationships. 
To enhance computational efficiency and minimize graph redundancy, we implement a heterogeneous spanning tree sampling method. Additionally, we devise a rotation-translation invariant geometric representation, ensuring broader applicability across diverse scenarios. Finally, we introduce Multipolygon-GNN, a novel model tailored to leverage the spatial and semantic heterogeneity inherent in the visibility graph.
Experiments on five real-world and synthetic datasets demonstrate its ability to capture informative representations for polygonal geometries. Code and data are available at \href{https://github.com/dyu62/PolyGNN}{$github.com/dyu62/PolyGNN$}.
\end{abstract}






\maketitle
\section{Introduction}
\begin{figure}
    % \centering
    \includegraphics[width=0.9\linewidth]{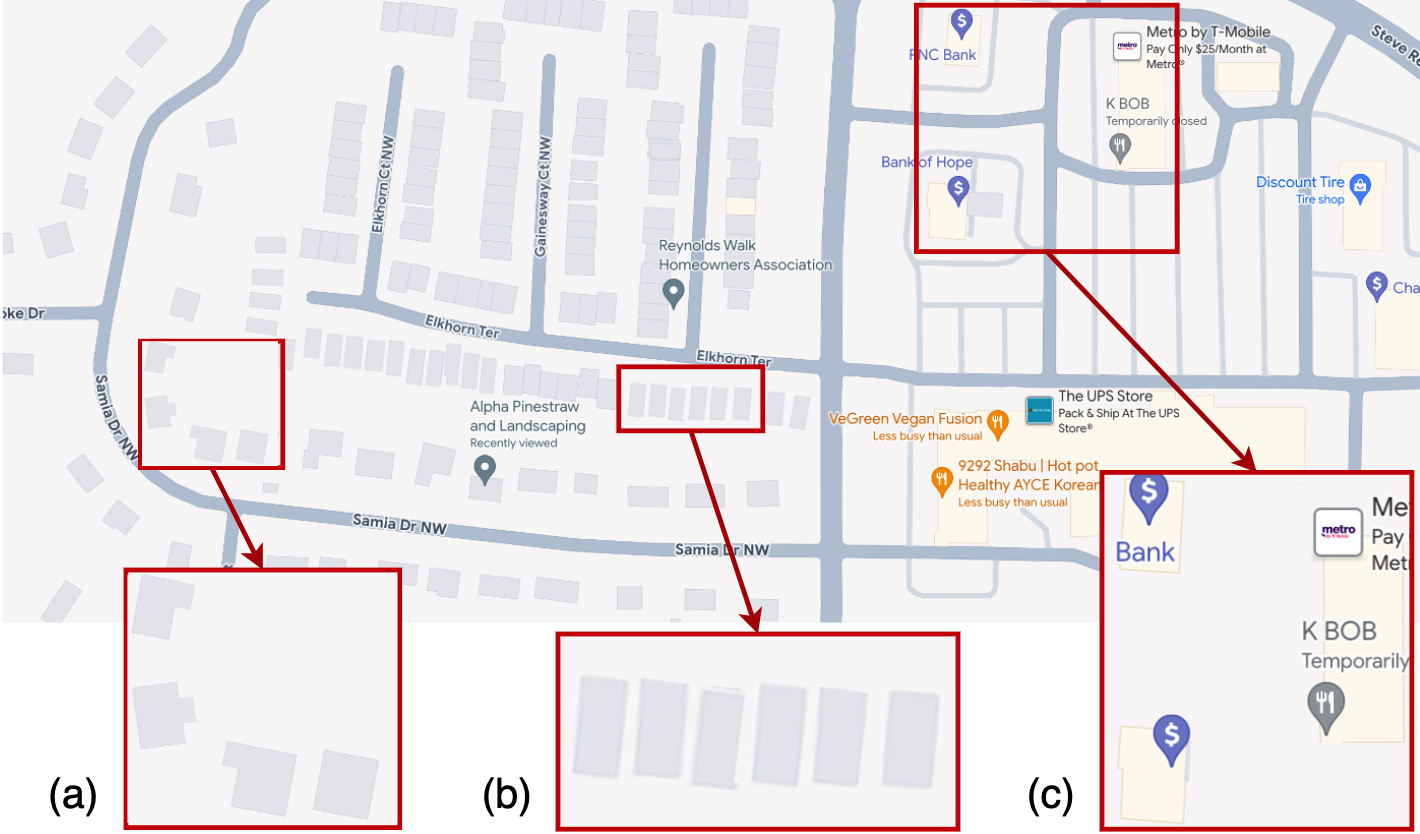}
    \caption{Illustrations of building multipolygon patterns for (a) houses, (b) townhouses, and (c) commercial buildings.}
    \label{fig:intro}
\end{figure}
Polygon representation learning refers to the process of capturing and encoding the essential features and characteristics of polygonal geometries, including those with or without holes and ranging from single to multipolygons. The learned embeddings can be leveraged directly for downstream applications such as urban planning \cite{zhong2020open,spangler2023calculating, mu2022computation}, shape coding \cite{li2020approximating, yan2021graph, zorzi2022polyworld}, building pattern recognition \cite{he2018recognition, yan2019graph, bei2019spatial}, cartographic generalization \cite{feng2019learning, wang2023hexagon, ma2020polysimp}, and geographic question answering (GeoQA) \cite{fabbri2020template, scheider2021geo,kefalidis2023benchmarking}.  
The accurate processing and interpretation of these polygonal shapes are crucial for achieving a nuanced understanding of our physical environment.

Current approaches in polygon representation learning primarily focus on individual polygon shapes \cite{mai2023towards}, often neglecting the critical inter-polygonal relationships. Such relationships are paramount in many scenarios but are inadequately addressed by existing methods. Current strategies either simplify polygons to points to consider inter-polygonal relations \cite{yan2021graph}, losing valuable shape information, or aggregate polygon representations without effectively capturing their interrelations \cite{van2019deep}. However, the geometries inside and across the polygons are both important. 
Figure \ref{fig:intro} showcases how the shapes and spatial distributions of buildings suggest their utility. Houses (Figure \ref{fig:intro} (a)) feature irregular shapes with a loose spatial arrangement along roads, aligning randomly with the road's contour. Townhouses (Figure \ref{fig:intro} (b)) are characterized by uniform shapes and sizes, arranged tightly and parallelly to optimize land use. Commercial buildings (Figure \ref{fig:intro} (c)) vary in shape and size based on business needs, often surrounding large customer parking spaces, with orientations parallel to the street for visibility. These diverse multipolygon patterns necessitate a holistic consideration of both individual shapes and their interrelations for effective representation learning, enabling the automated identification of area functionality.

Developing a machine learning model that effectively captures both the intricate details of individual polygons and their relationships presents significant challenges.  The first hurdle is devising a data structure capable of conserving geometric details across both the inner-polygonal relationships (shape details of individual polygons) and the inter-polygonal relationships (broader spatial dynamics between different polygons).  This requires an approach that unifies details with the overarching spatial context, ensuring no geometric information is lost. Furthermore, the intricate pairwise relationships between multiple polygons introduce quadratic complexity, which further calls for the efficiency of the representation learning method. Additionally, a pivotal aspect of polygon representation learning is the generalizability of the derived representations. Traditionally, multipolygons are represented by an ordered set of coordinates, which inherently encode specific orientations and positions. This method, however, does not account for rotation and translation invariance, limiting the ability to consistently interpret polygonal geometries irrespective of their spatial orientation or position. Finally, multipolygon inherently encompasses hierarchy notions which are crucial yet not well explored. In particular, a large multipolygon may consist of several small multipolygons, each exhibiting distinct patterns. For instance, rows of townhouses may collectively form a larger community structure. Recognizing and effectively modeling these hierarchical relationships is crucial for a comprehensive understanding of multipolygon configurations. Consequently, advanced models are needed to characterize these hierarchically organized patterns.      

To tackle these challenges, we propose PolygonGNN, a new framework that learns the representation of multi-polygons that ensures the preservation of paramount geometric information and characteristics of polygons.
Firstly, to concisely encapsulate both shape details and inter-polygonal relationships, we model a heterogeneous visibility graph. Here, graphs adeptly represent polygon shapes through vertices and edges, while visibility connections capture spatial relations between polygons.
% This heterogeneous visibility graph serves as an efficient data structure that captures both intricate shape details and inter-polygonal relationships, ensuring the preservation of vital topological information essential for accurate representation. 
Secondly, to reduce the redundancy in the heterogeneous visibility graph and boost training efficiency, we develop a heterogeneous spanning tree sampling strategy that selectively samples visibility edges within a multipolygon. Moreover, the randomness of edge sampling also enriches the training dataset with various graph instances.
% This approach selectively samples visibility edges within a multipolygon, creating a spanning tree where each polygon is represented as a supernode. This approach guarantees the graph remains strongly connected and the conversion process from multipolygon to graph retains its invertibility, thus preserving topological integrity. Moreover, the randomness of edge sampling also enriches the training dataset with various graph instances.
To preserve the multipolygon's geometric information with rotation and translation invariance, we consider the two-hop path set within the graph and propose a heterogeneous geometric representation for the nodes in the heterogeneous visibility graph. This heterogeneous geometric representation is proven to encapsulate the complete spatial and semantic information present in the graph, ensuring the model's invariance to rotation and translation. 
To effectively learn hierarchical patterns of multipolygons, we develop Multipolygon-GNN, which is a new graph neural network that stacks multiple layers of heterogeneous geometric message passing operators that aggregate multipolygon patterns in different granularities.
% The graph neural network stacks multiple layers to learn multi-hop relationships within a multipolygon, which inherently introduces a hierarchical structure tailored for analyzing multipolygon configurations.  
% Lastly, to effectively utilize the spatial representation tuple and accommodate the graph's heterogeneity, we propose a tailored graph neural network (GNN),  Multipolygon-GNN. This model processes messages according to four distinct types of interactions based on the two-hop paths, employing specialized neural networks for each interaction type.  This design adeptly captures the semantic heterogeneity of the visibility graph. Furthermore, we rigorously demonstrate that our proposed model possesses significant expressive power, enabling it to distinctively identify and differentiate between diverse input graphs. 

Our contributions are summarized as follows:
\begin{itemize}
[leftmargin=*]
  \item {We propose an invertible process for transforming multipolygons to heterogeneous visibility graphs to integrate polygon shape details with inter-polygonal relationships.}
  \item We propose heterogeneous spanning tree sampling that improves training efficiency without sacrificing geometric fidelity.
  \item {We develop a lossless rotation-translation-invariant heterogeneous geometric representation for the visibility graphs. } 
  \item We propose Multipolygon-GNN that capitalizes on the spatial and semantic heterogeneity of the visibility graphs.
  % \item We conduct comprehensive evaluations on diverse real-world and synthetic datasets, underscoring our methodology's effectiveness over existing polygon representation learning methods. 
\end{itemize}

\section{Related Work}
\subsection{Polygon Representation Learning}

Recently, we have seen many research advancements in representation learning for polygonal geometries \cite{jiang2022weakly,jiang2021weakly,he2022quantifying}. Existing approaches primarily fall into three categories: 1) traditional feature engineering methods, 2) polygon shape encoding methods, and 3) multipolygon representation learning. Each of these methodologies, while contributing valuable insights and capabilities, faces its challenges: 
Traditional feature engineering methods \cite{pham2010fast, yan2019graph, he2018recognition} transform polygon shapes into predefined shape descriptors for neural network processing. However, these descriptors often oversimplify, failing to capture the full spectrum of shape information and requiring significant domain knowledge for their design. Their inability to handle the variability and complexity of polygonal shapes limits their generalizability. 
Polygon shape encoding methods mainly focus on encoding simple polygons without holes \cite{van2019deep, mai2023towards, yan2021graph}. Veer et al. \cite{van2019deep} proposed two encoders for simple polygons based on 1D CNN and bidirectional LSTM, taking the exterior coordinate matrix as input. Mai et al. \cite{mai2023towards} used a 1D ResNet architecture with circular padding for loop origin invariance. In \cite{yan2021graph}, authors treated the polygon exterior as a graph and used a graph autoencoder for embedding. These methods have shown effectiveness on tasks like shape classification and retrieval. These methods though useful for certain types of analyses, are not ideally suited for multipolygon representation learning where capturing the intricate topological relationships between polygonal geometries is needed.  

For multipolygon representation learning, Jiang et al. \cite{jiang2019ddsl} proposed to use Non-Uniform Fourier Transform (NUFT) to first transform polygons and 3D meshes into the Fourier domain before feature extraction with CNNs. On this basis, Mai et al. \cite{mai2023towards} developed NUFTspec to directly learn embeddings in the spectral domain instead of converting back to the spatial domain. NUFTspec has shown promise in predicting spatial relationships such as topology and cardinal direction relationships, which are useful in geographic question answering \cite{hamzei2022translating, kuhn2021semantics, mai2021geographic}. Despite these advances, there remains a gap in effectively handling the intricacies of multipolygon geometries, especially in capturing both the inter-polygonal relationships and the nuances of individual polygon shapes.

% More recent works have explored encoding polygons with complex geometries like holes and multipolygons. 

\subsection{Graph Neural Networks for Multipolygons}
Graphs are inherently suited to model the complex spatial relationships and interactions within spatial systems \cite{yu2022deep,zhang2022deep,zhang2023deep,zhang2021representation}, making them ideal for multipolygon analysis \cite{cosma2023geometric, zhao2022extracting, gueze2023floor, antonietti2024agglomeration}.  He et al. \cite{he2018recognition} develop a graph-centric method that utilizes graph representations and combines graph partitioning with random forests to identify and classify diverse building patterns.  Yan et al.  \cite{yan2019graph, bei2019spatial} utilize graph convolutional neural networks to classify urban building patterns using spatial vector data, which is designed to distinguish regular and irregular group patterns to enhance map generalization and urban planning.  
However, while these homogeneous graph models are effective in modeling spatial relationships among multiple polygons, they are limited in handling diverse non-homogeneous data types, such as the distinct types of entities and interactions within multipolygons.  Heterogeneous Graph Neural Networks (HGNNs) introduce advanced mechanisms to manage this diversity \cite{gillsjo2023polygon}.  
HAN \cite{han} leverages node-level attention and semantic-level attention to learn the importance of nodes and meta-paths, respectively. HGT \cite{hu2020heterogeneous} uses meta relation to decompose the interaction and transform matrices to model heterogeneity. Yet, a bespoke HGNN solution specifically designed for the multipolygon challenge is notably absent from the literature, indicating a clear avenue for innovation.

% Generally, HGNNs are divided into three primary architectures: convolution-based, autoencoder-based, and adversarial-based models, each offering unique approaches to preserve and learn from the heterogeneity of data.

% In addition, metapath-free models like GTN \cite{gtn} and HetSANN \cite{hetsann} learn node embeddings by aggregating neighbors with relation importance. These methods are effective at feature learning but have high complexity and space consumption. Overall, convolution-based models such as HAN and HGT achieve a balance between effective aggregation and computational efficiency compared to more complex or unsupervised alternatives and remain the best choice in heterogeneous graph representation learning.

\section{Problem Formulation}
In this section, we first introduce the definitions of polygon and multipolygon, and then we formalize the problem of multipolygon representation learning.
\begin{defn} Polygon. A polygon $p_i$ is defined as a tuple $(\mathbf{B_i},h_i=\{\mathbf{H}_{i,j}\}_{j=1}^{N_{h_i}})$, where $\mathbf{B_i}\in \mathbb{R}^{N_{b_i}\times 2}$ represents the coordinates of the vertices on the exterior boundary of $p_i$, arranged in a \textit{counterclockwise} direction.
% Vertices of polygons do not overlap. 
% Vertices of holes follow a similar logic. 
$h_i$ denotes the set of holes within $p_i$, with $\mathbf{H}_{i,j}\in \mathbb{R}^{N_{h_{i,j}}\times 2}$ indicating the coordinates of vertices for the $j$-th hole, oriented in a \textit{clockwise} direction. 
% $j\in [1, N_{h_i}]$,
$N_{h_i}$, $N_{h_{i,j}}$ and $N_{b_i}$ denote the number of holes in $p_i$, the number of vertices on the boundary of $j$-th hole and the number of vertices on $p_i$'s exterior boundary, respectively. Both $\mathbf{B_i}$ and $\mathbf{H}_{i,j}$ create closed shapes which are defined as polygon parts. Altering the starting vertex in $\mathbf{B_i}$ and $\mathbf{H}_{i,j}$ does not change the structure of the polygon.
% Additionally, a polygon remains unchanged under transformations such as translation and rotation applied to its coordinates.
\end{defn}
\begin{defn} Multipolygon. A multipolygon $q$ is defined as a collection of polygons $\{p_{i}\}_{i=1}^{N_q}$, where $N_q$ denotes the number of polygons. A multipolygon reduces to a single polygon if $N_q=1$. 
% The polygons within a multipolygon do not overlap each other. 
\end{defn}
As depicted in Figure \ref{fig:intro}, individual polygons may possess similar shapes across various scenarios, yet when aggregated into multipolygons, they exhibit a wide array of patterns. This observation highlights the necessity of evaluating multiple single polygons collectively to accurately identify group patterns, which is fundamental to understanding the concept of a multipolygon. 
This paper aims to convert a multipolygon into a vector representation, denoted as $q \rightarrow q_v$, where $q_v \in R^d$ and $d$ represents the dimension of the vector. The learned representation should be able to be used to discriminate different multipolygons they represent and maintain necessary geometric information and properties in order to be used in downstream applications.
% , ensuring that the resultant representation can be effectively incorporated into a variety of downstream applications.

Representation learning of multipolygon is an underexplored yet highly challenging problem due to the key technical hurdles: 1) Difficulty in unifying the modeling of the inner polygonal structure and the relationships between multiple single polygons. 2) Difficulty in handling the scalability issue.3) Difficulty in learning a unique representation for different input multipolygon. 4) Difficulty in learning the heterogeneous relationships within a multipolygon.
\section{Methodology}
\begin{figure*}[htb]
  \centering
  \includegraphics[width=\textwidth]{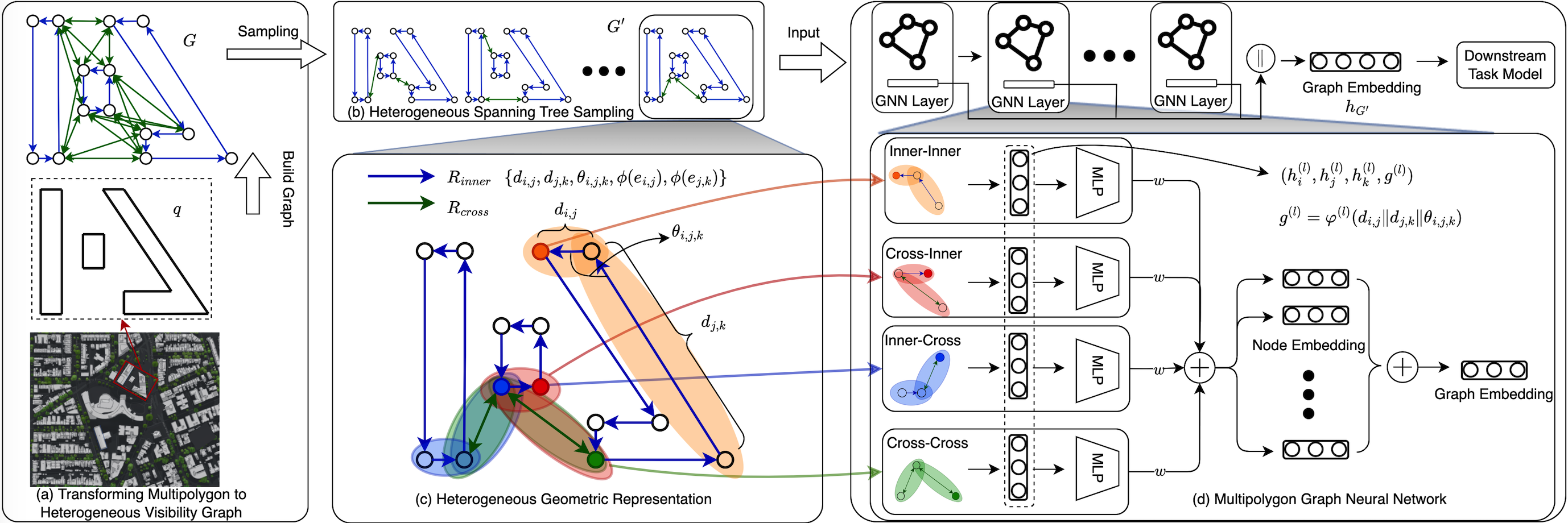}
  \caption{Illustration of the proposed framework.}
  \label{fig:arch}
\end{figure*}
To learn distinct representations for multipolygons by addressing the aforementioned challenges, we propose the PolygonGNN framework as shown in Figure \ref{fig:arch}. First, in Figure \ref{fig:arch}-(a), to unify the characterization of the inter-polygonal and inner-polygonal relationships in multipolygon data, we propose a new transformation that turns a multipolygon into a heterogeneous visibility graph, as elaborated in Section \ref{sec:graph build}. This process is proven to be invertible, which maintains necessary information in the multipolygon while converting it to a well-structured data format. In Figure \ref{fig:arch}-(b), to reduce the redundancy in the visibility graph and solve the scalability issue (introduced in Section \ref{sec:sampling}), we propose to sample out succinct graphs that are still sufficient to preserve multipolygon information. (Section \ref{sec:sampling}). Figure \ref{fig:arch}-(c) illustrates our five-tuple heterogeneous geometric representation for each two-hop path. This representation is designed to transform geometry into vector form while still preserving all the information of the visibility graph and achieving rotation and translation invariance (Section \ref{sec:spatial rep}). The representation further eliminates the redundancy in the heterogeneous visibility graph. Finally, as shown in Figure \ref{fig:arch}-(d), to characterize the hierarchical patterns of the multipolygon, multiple layers of heterogeneous two-hop message-passing mechanism are proposed in Section \ref{sec:hgnn}.  It learns the contexts of nodes hierarchically without losing geometric information and properties. We demonstrate that this heterogeneous graph neural network is capable of distinguishing between various input graphs. 

\subsection{Transforming Multipolygon to Heterogeneous Visibility Graph } \label{sec:graph build}
A multipolygon is characterized not only by the shape of its constituent parts but also by the spatial relationships between these parts. Developing a comprehensive representation of multipolygons necessitates a unified data structure capable of encapsulating both shape geometry and spatial interconnections. The geometry of each part is delineated by an ordered set of coordinates, which lends itself to graph representation. Each graph node represents a polygon vertex, with its coordinates as node features, and edges delineate the shape of the part.
To model the spatial relationships inherent in multipolygons, we calculate the visibility relationships among the parts of the multipolygon, transforming abstract coordinates into pairwise spatial relationships. This introduces a layer of rich semantics essential for spatial applications in the real world.
To harmonize these dual aspects of multipolygon representation, we propose the concept of a heterogeneous visibility graph, specifically tailored for multipolygon contexts. This graph incorporates two distinct types of edges: inner edges that define the shapes of individual parts and visibility edges that connect separate parts, modeling their spatial relationships. 

As shown in Figure \ref{fig:arch}-(a), given a multipolygon $q = \{p_{i}\}_{i=1}^{N_q}$,  $p_{i}=(\mathbf{B_i},h_i=\{\mathbf{H}_{i,j}\}_{j=1}^{N_{h_i}})$, the heterogeneous visibility graph is defined as $G(V, E, X, \phi)$.
Here, each node $v_i \in V$ represents a polygon vertex, the node set $V$ comprises vertices recorded in $\mathbf{B_i}$ and $\mathbf{H}_{i,j}$, with $|V| = {N_{b_i}+\sum_{j=1}^{N_{h_i}}N_{h_{i,j}}}$. 
Each edge $e_{i,j} \in E$ is associated with a specific edge type. $X$ represents the node attributes, which are set initially using their respective coordinates.
Let $\phi: E\rightarrow \mathcal{R}$ represent an edge type mapping function, where $\mathcal{R}=\{R_{inner}, R_{cross}\}$ denotes the set of edge types. 
Edges of the type $R_{inner}$ are directed \textit{inner edges} which trace the shape of each closed polygon part. The creation of inner edges involves a sequential connection of vertices. Specifically, an inner edge is formed between two consecutive nodes as recorded in $\mathbf{B_i}$ or $\mathbf{H}_{i,j}$. Notably, there is also an inner edge connecting the last and first vertices in $\mathbf{B_i}$ and $\mathbf{H}_{i,j}$, respectively, to form a closed shape.
Edges of the type $R_{cross}$ are undirected \textit{visibility edges} that connect separated parts. Two nodes in $G$ are considered visible to each other if the line segment connecting them does not intersect any polygon boundaries. 
% Visibility edges are established only between pairs of nodes that are visible to each other. 
% Visibility between two parts is defined if there exist nodes in each part that are visible to each other. A node is visible to a part if there is at least one node in that part visible to the given node. 
For the visibility edges, we iterate over $V$ and build edges between pairs of nodes that are visible to each other.  
This process ensures that the graph not only represents the physical structure of the multipolygons but also maintains connectivity, enabling a holistic understanding of the spatial network. In addition, we have the following theorem which shows that the transformation from a multipolygon to a heterogeneous visibility graph is information lossless.
\begin{thm}\label{thm1} Let $q$ be a multipolygon and $G(V,E, X, \phi)$ be the heterogeneous visibility graph derived from $q$, the transformation to the graph is invertible. 
\end{thm}
% \begin{proof}
%     The detailed proof is in Appendix \ref{appendix:thm1}.
% \end{proof}
The detailed proof is in Appendix \ref{appendix:thm1}. We prove it by using the graph information to rebuild an equivalent multipolygon.  Essentially, we have successfully translated the challenge of representing multipolygons into a graph representation problem without any loss of crucial data.

\subsection{Heterogeneous Spanning Tree Sampling}\label{sec:sampling}
The concept of heterogeneous spanning tree sampling addresses the redundancy in the heterogeneous visibility graph. By leveraging the characteristics of heterogeneous visibility graphs, we developed a linear complexity sampling strategy.
Key to this approach is the understanding that visibility edges serve to connect separated parts of a polygon. To ensure effective message exchange between different parts, it suffices to maintain at least one path connecting each distinct part. This requirement can be effectively handled by solving a spanning tree problem.
As shown in Figure \ref{fig:arch}-(b), we conceptualize different polygon parts as supernodes interconnected by visibility edges. By randomly sampling visibility edges, we construct a spanning tree $\omega \in \Omega$ that guarantees the connectivity of these supernodes. 
% This approach also guarantees the conversion process from multipolygon to graph retains its invertibility, thus preserving topological integrity. 

Given a heterogeneous visibility graph $G(V, E, X, \phi)$, we categorize its edge set into two types: $E=E_{inner}\cup E_{cross}$. Here $E'_{cross,\omega}$ represents the subset of selected visibility edges forming a spanning tree $\omega$. Consequently, the newly derived graphs are represented as $\{G(V, E_{inner}\cup E'_{cross,\omega}, \phi)| \omega \in \Omega\}$  We also establish that the sampled graphs can be inversely transformed back to their corresponding original multipolygon, thus ensuring that no information is lost during the sampling process.
\begin{co}\label{thm-aug}
For a multipolygon $q$ and its associated heterogeneous visibility graph $G(V,E, \phi)$, each graph  $G' \in \{G(V, E_{inner}\cup E'_{cross,\omega}, \phi)|\omega \in \Omega \}$  uniquely identifies the original multipolygon $q$.
\end{co}
The detailed proof is in Appendix \ref{appendix:thm-aug}. In addition, the sampling not only reduces computational complexity but also acts as a form of data augmentation, generating more training graphs from which the model can benefit. Our experiments demonstrate an improvement in performance attributable to this augmentation.

\subsection{Heterogeneous Geometric Representation}\label{sec:spatial rep}
In a sampled heterogeneous visibility graph $G'$, the spatial information is encapsulated by the relative positions of polygon parts and the specific shape of each part, which are defined by the coordinates of the polygon vertices. 
To effectively distinguish between two different polygons, we need to account for these coordinates to maintain spatial awareness during the update of node embeddings. However, it is crucial to recognize that polygonal structures remain unchanged under transformations like translation and rotation. Therefore, achieving rotation and translation invariance becomes a crucial goal, which is not inherently provided by the coordinates themselves. To address this, one approach is to transform the original coordinates into distances between nodes, serving as features. This method can achieve the desired invariance to translation and rotation. However, relying solely on distances might not be sufficient to preserve the complete spatial information. For instance, if the edge connecting nodes $v_i$ and $v_j$ rotates around $v_i$, the distance between these nodes remains constant, hence not reflecting the change in the polygon's structure. This limitation indicates that distance-only features are insufficient for a model to fully capture the spatial dynamics of the polygon. Moreover, the heterogeneity of the graph assigns different semantics according to whether we are considering positions within a single polygon or the positions between multiple polygons. To overcome this limitation, we propose a five-tuple heterogeneous geometric representation that is proven to comprehensively represent the heterogeneous visibility graph converted from a multipolygon.
Specifically, consider $\Pi^{i}_{2}$, the set of all \textit{two-hop paths} converging to node $v_{i}$. In this context, $\pi_{i,j,k} \in \Pi^{i}_{2}$ indicates a \textit{two-hop path} $v_{i}\leftarrow v_{j}\leftarrow v_{k}$.  The heterogeneous visibility graph $G'(V, E, X, \phi)$ can be expressed as a collection of tuples $s(\pi_{i,j,k})$:
% \begin{equation}\label{eq:path3}
%     s(\pi_{i,j,k})=(d_{i,j},d_{j,k},\theta_{i,j,k}),
% \end{equation}
\begin{align}
    &G'=\{s(\pi_{i,j,k})|\pi_{i,j,k} \in \Pi^{i}_{2}, v_i \in V\}, \label{eq:path3} \\
    &s(\pi_{i,j,k})=(d_{i,j},d_{j,k},\theta_{i,j,k}, \phi(e_{i,j}),\phi(e_{j,k}) ) \nonumber
\end{align}
where $d_{i,j}$ denotes the distance between node $v_i$ and $v_j$, $d_{j,k}$ is the distance between node $v_j$ and $v_k$, $\theta_{i,j,k} \in [-\pi, \pi)$ is the angle at $v_j$ formed by the three nodes. $\phi(e_{i,j}),\phi(e_{j,k})$ are the types of the edges forming the path, as shown in Figure \ref{fig:arch}-(c).
% where $d_{ij}=||\mathbf{P}_{ij}||_{2}$ and $ \theta_{i,j,k}=arccos(<\frac{\mathbf{P}_{ij}}{d_{ij}},\frac{\mathbf{P}_{jk}}{d_{jk}}>)$. $||\mathbf{P}_{ij}||_{2}=\sqrt{\mathbf{P}_{ij}\cdot \mathbf{P}_{ij}}$ where $\mathbf{P}_{ij} = \mathbf{P}_{i}-\mathbf{P}_{j}$ and $\mathbf{P}_{i}$, $\mathbf{P}_{j}$ represent the Cartesian coordinates of node $v_{i}$, $v_{j}$.
Importantly, the representation is invariant under rotation and translation transformations, ensuring that the structural integrity of the graph is maintained regardless of its orientation or position.
We further affirm that this tuple format encapsulates all the heterogeneous spatial information of the graph. In essence, utilizing the tuple representation of the heterogeneous visibility graph, as detailed in Equation \ref{eq:path3}, enables us to reconstruct a graph that is functionally equivalent to the original. 
\begin{lem}\label{lem} Given a tuple $s(\pi_{i,j,k})$ and the positions of $v_{i}$ and $v_{j}$ within a two-hop path $\pi_{i,j,k}$, we can determine the position of the third node $v_{k}$, as well as the types of edges that constitute the path.
\end{lem}
The detailed proof is in Appendix \ref{appendix:lem}. The lemma indicates that the geometric configuration of a two-hop path $\pi_{i,j,k}$ is rigidly determined by the tuple $s(\pi_{i,j,k})$, akin to a rigid body. Concurrently, it establishes the types of edges comprising the path, thereby clarifying whether any two nodes originate from two polygon parts. 
\begin{thm}\label{lem-thm} 
Given the tuple representation of a heterogeneous visibility graph $G'$, as articulated in Equation \ref{eq:path3}, we can reconstruct a graph that is equivalent to $G'$.
\end{thm}
The detailed proof is in Appendix \ref{appendix:lem-thm}. The foundational concept of Theorem \ref{lem-thm} is that all parts within a multipolygon are interconnected via undirected edges. So we can iteratively expand the rigid structure, thereby reconstructing all nodes and their interconnectivity to form an equivalent heterogeneous visibility graph. 
 
% This model processes messages according to four distinct types of interactions based on the two-hop paths, employing specialized neural networks for each interaction type.  This design adeptly captures the semantic heterogeneity of the visibility graph. 
\subsection{Multipolygon Graph Neural Network}\label{sec:hgnn}
To effectively learn the hierarchical representation of a multipolygon, we propose Multipolygon-GNN with a focus on utilizing different models to learn the different interactions in the heterogeneous visibility graph. In each layer, we utilize a two-hop message-passing scheme to update node embeddings leveraging the five-tuple heterogeneous geometric representation outlined in Section \ref{sec:spatial rep}.
As shown in Figure \ref{fig:arch}-(d), a two-hop path $\pi_{i,j,k}$ can connect nodes within the same polygon part or between different parts, through various edge types. The flow of information from one polygon part to another is critical in learning the interrelation among multiple polygons, while inner-part information flow enhances the understanding of local context and spatial relationships within a single polygon. We categorize possible path types based on the edge types involved as four categories. Let $\Phi$ be the two-hop path type mapping function,
\begin{equation}
    \Phi(\pi_{i, j, k}) \in \{R_{inner}, R_{cross}\} \times \{R_{inner}, R_{cross}\}
\end{equation}
To effectively utilize the graph's heterogeneity and distinguish between different message sources, we propose a heterogeneous function for learning the message based on the path type. Let $\Psi^{(l, \Phi(\pi_{i, j, k}))}$ be an MLP model for path type $\Phi(\pi_{i, j, k})$ at layer $l$, the message $m^{(l)}(\pi_{i, j, k})$ from path $\pi_{i,j,k}$ can be formulated as follows:
\begin{equation}
m^{(l)}(\pi_{i, j, k})=w^{(\Phi(\pi_{i, j, k}))}\Psi^{(l, \Phi(\pi_{i, j, k}))} (h_i^{(l)}, h_j^{(l)}, h_k^{(l)}, g^{(l)}),
\end{equation}
where $w^{(\Phi(\pi_{i, j, k}))}$ is the weight for path type $\Phi(\pi_{i, j, k})$, $g^{(l)}=\varphi^{(l)}(d_{i, j} \Vert d_{j, k} \Vert  \theta_{i, j, k})$ is the geo-embedding calculated by an MLP function $\varphi^{(l)}$, and $\Vert$ denotes the concatenation operation.
% [A GNN recursively updates each node’s feature vector through message-passing. 
We initialize node embeddings to zeroes. For a node $v_i$, let $h_i^{(l+1)}$ represent its updated embedding in the $l$-th layer. The node embedding update is formulated as follows:
\begin{equation}
    h_i^{(l+1)}=  \sum \{ m^{(l)}(\pi_{i, j, k})| \pi_{i,j,k} \in \Pi_2^i\}.
\end{equation}
To maximize discriminative power, the embeddings of all nodes are summed to form a graph embedding, and the graph embeddings from all layers are concatenated as the final graph representation $h_G$ for downstream tasks:
\begin{equation}
    h_{G'}=\big\Vert_{l=1}^L \left(\sum\nolimits_{i=1}^{|V|} h_i^{(l)}\right),
\end{equation}
where $L$ is the number of GNN layers. 

Our two-hop message-passing approach, incorporating distances and angles, achieves rotation and translation invariance, while retaining the ability to distinguish different graphs. Assuming the distance between any two nodes is bounded within a range, we demonstrate that our method can aggregate complete graph information with arbitrary precision:
\begin{thm}\label{thm-haus}
Suppose $f: \mathcal{S} \rightarrow \mathbb{R}$ be a continuous set function with respect to the Hausdorff distance $d_H(\cdot,\cdot)$. Let $S \in \mathcal{S}$ be the set of all two-hop paths of a heterogeneous visibility graph $G$,  $S=\{s(\pi_{i,j,k}) | v_i \in V\}$, $\forall \epsilon>0, \exists K \in \mathbb{Z}$, such that for any $S \in \mathcal{S}$, 
\begin{equation}
    |f(S)-\zeta(f'(S))|<\epsilon,
\end{equation}
where $\zeta$ is a continuous function, and $f'(S)\in \mathbb{R}^K$ is the output of our proposed method.
\end{thm}
The detailed proof is in Appendix \ref{appendix:thm-haus}. Similar to PointNet, in the worst case, our method divides the space into small granules. With a sufficiently large output dimension, our method maps each input into a unique granule.

\section{Experiment}
We first introduce the experimental settings. The implementation details including hyperparameter tuning can be found in Appendix \ref{appendix:implementation}. Then we present the performance of PolygonGNN and comparison approaches across five distinct datasets for graph classification.  Then we perform ablation studies to show the contributions of the sampling strategy and Multipolygon-GNN. We also include analyses on efficiency, hyperparameter sensitivity, embedding visualizations, and heterogeneous interaction weights, alongside a case study to identify the strengths and limitations of our framework.

\subsection{Experiment Setting}
\subsubsection{Datasets}
Our experiments span five datasets featuring varied types of spatial polygons, detailed as follows:
\textbf{ MNIST-P-2:} This dataset comprises 10,000 samples of two-digit multipolygons, each formed by combining two random digits from the MNIST-P dataset \cite{jiang2019ddsl}. The MNIST-P, a polygonal adaptation of the MNIST dataset \cite{lecun1998gradient}, represents each digit as a polygonal shape derived by tracing the original images' contours. 
% For this study, we adjust the polygon vertices to mimic the appearance of realistic buildings. 
The label for each two-digit sample merges the individual digits' labels, sequenced from left to right, resulting in a classification task across 90 classes (digits 10-99).
\textbf{ Building-2-R:} Containing 3,469 two-building multipolygons, each entry in this dataset pairs two buildings from the OpenStreetMap (OSM) building dataset \cite{yan2021graph}. A building's label reflects its shape, categorized into one of ten standard alphabetic shapes (H, I, E, Y, T, F, U, L, Z, O). Buildings are coupled with their closest neighbor on the OSM map, and the label for each pair accounts for their relative positioning.
\textbf{ Building-2-C:} This dataset includes 5,000 samples of two-building multipolygons. Buildings are resized to fit within a 1x1 area and randomly paired, with the composite label reflecting the sequence of their individual labels from left to right.
\textbf{ Building-S:} Features 5,000 single-building polygons for shape classification, focusing on identifying the buildings' geometric shapes.
\textbf{ DBSR-cplx46K:} Introduced by \cite{mai2023towards}, this extensive dataset contains 46,567 complex polygonal geometry samples, each comprising two polygons. The research objective is to predict spatial relations, specifically classifying whether one of the two polygons partially contains the other.

\subsubsection{Comparison Method}
In evaluating the proposed framework, we benchmark it against several leading-edge baselines renowned for their prowess in polygon encoding:
\textbf{ResNet1D:} This model adapts the 1D variant of the Residual Network (ResNet) architecture, incorporating circular padding to effectively encode the exterior vertices of polygons. \textbf{VeerCNN:} A Convolutional Neural Network (CNN) designed for 1D inputs, VeerCNN employs zero padding and concludes with global average pooling. \textbf{ NUFT-DDSL:} A spatial domain polygon encoder that uses NUFT features and the DDSL model. \textbf{ NUFT-IFFT:} A spatial domain polygon encoder that utilizes NUFT features and the inverse Fast Fourier transformation (IFFT).
Additionally, to underscore the efficacy of our proposed multipolygon graph neural network, we include comparisons with two heterogeneous GNNs:
\textbf{HAN:} A GNN implements an attention mechanism to manage the diverse relationships within heterogeneous graphs.
\textbf{HGT:} A GNN adapts the transformer architecture to leverage node and edge type-specific attention mechanisms to capture complex patterns.

% \subsubsection{Implementation Details}
% We leverage MLPs for the spatial feature extractor $\zeta_{SE}$ and the decoder $SD$. For each dataset, we randomly split it into 60\%, 20\%, and 20\% for training, validation, and testing respectively. All the Deep learning models are trained for a maximum of 200 epochs using an early stop scheme. Adam optimizer with a learning rate of 0.001 is used. 

\begin{table*}
\small
  \caption{The performance of the proposed model (including ablation variants) and the comparison methods. The best results are in bold. }
  \centering
  \begin{tabular}{c|c|cccc|cccc}
    \toprule
    Dataset & Metric & ResNet1D   & VeerCNN   & NUFT-DDSL  & NUFT-IFFT & HAN     & HGT & PolygonGNN w/o S &PolygonGNN\\
    \midrule
    MNIST-P-2 & Acc   & 0.794 $\pm$.012&   {0.667$\pm$.019} & {0.559$\pm$.014} & {0.357$\pm$.029} & {0.865$\pm$.013} & {0.872$\pm$.016} &0.880$\pm$.009    &\textbf{0.897$\pm$.004}\\
    & Prec   & 0.810$\pm$.018 &   {0.709$\pm$.017} & {0.593$\pm$.013} & {0.391$\pm$.027} & {0.871$\pm$.012} & {0.877$\pm$.010} &0.885$\pm$.012 &\textbf{0.901$\pm$.010}\\
    & F1   & 0.794$\pm$.012 &   {0.667$\pm$.018} & {0.561$\pm$.014} & {0.357$\pm$.028} & {0.865$\pm$.012} & {0.872$\pm$.012} &0.879$\pm$.010 &\textbf{0.897$\pm$.007}\\
    & AUC  & 0.995$\pm$.001 &   {0.986$\pm$.008} & {0.964$\pm$.005} & {0.908$\pm$.010} & {0.996$\pm$.001} & {0.996$\pm$.001} &{0.996$\pm$.001} &\textbf{0.997$\pm$.000}\\
    \hline
    Building-2-C & Acc 
    & 0.146$\pm$.020 &   {0.121$\pm$.005} & {0.088$\pm$.023} & {0.059$\pm$.031} & {0.318$\pm$.027} & {0.347$\pm$.023} &0.536$\pm$.013 & \textbf{0.537$\pm$.025}\\
    & Prec   & 0.175$\pm$.026 &   {0.125$\pm$.007} & {0.108$\pm$.030} & {0.072$\pm$.034} & {0.331$\pm$.045} & {0.367$\pm$.041} &0.568$\pm$.014 &\textbf{0.578$\pm$.026}\\
    & F1   & 0.145$\pm$.021 &   {0.111$\pm$.006} & {0.086$\pm$.024} & {0.060$\pm$.038} & {0.310$\pm$.033} & {0.339$\pm$.029} &0.530$\pm$.013 &\textbf{0.537$\pm$.025}\\
    & AUC  & 0.860$\pm$.055 &   {0.836$\pm$.010} & {0.738$\pm$.051} & {0.703$\pm$.055} & {0.932$\pm$.014} & {0.944$\pm$.015} &0.984$\pm$.003 &\textbf{0.985$\pm$.008}\\
    \hline
    Building-2-R 
    & Acc   & 0.464$\pm$.014 &   {0.372$\pm$.060} & {0.244$\pm$.028} & {0.244$\pm$.013} & {0.599$\pm$.079} & {0.637$\pm$.071} &0.659$\pm$.020 &\textbf{0.663$\pm$.011}\\
    & Prec   & 0.505$\pm$.014 &   {0.375$\pm$.111} & {0.278$\pm$.029} & {0.287$\pm$.015} & {0.623$\pm$.083} & {0.651$\pm$.077} &0.679$\pm$.020 &\textbf{0.696$\pm$.021}\\
    & F1   & 0.451$\pm$.014 &   {0.352$\pm$.079} & {0.223$\pm$.023} & {0.229$\pm$.025} & {0.585$\pm$.080} & {0.625$\pm$.076} &0.642$\pm$.019 &\textbf{0.646$\pm$.015}\\
    & AUC  & 0.855$\pm$.005 &   {0.843$\pm$.025} & {0.736$\pm$.008} & {0.710$\pm$.009} & {0.917$\pm$.039} & {0.946$\pm$.026} &\textbf{0.969$\pm$.010} &{0.964$\pm$.008}\\
    \hline
    Building-S 
    & Acc  & 0.749$\pm$.016 &   {0.643$\pm$.059} & {0.847$\pm$.005} & {0.814$\pm$.002} & {0.898$\pm$.007} & {0.950$\pm$.004} &{0.984$\pm$.009} &\textbf{0.984$\pm$.007}\\
    & Prec   & 0.773$\pm$.015 &   {0.658$\pm$.073} & {0.861$\pm$.005} & {0.846$\pm$.001} & {0.901$\pm$.006} & {0.951$\pm$.004} &{0.984$\pm$.009} &\textbf{0.983$\pm$.007}\\
    & F1   & 0.748$\pm$.018 &   {0.644$\pm$.055} & {0.847$\pm$.006} & {0.817$\pm$.001} & {0.898$\pm$.006} & {0.950$\pm$.004} &{0.984$\pm$.009} &\textbf{0.984$\pm$.007}\\
    & AUC  & 0.954$\pm$.005 &   {0.934$\pm$.021} & {0.986$\pm$.001} & {0.984$\pm$.000} & {0.992$\pm$.000} & {0.998$\pm$.000} &\textbf{0.999$\pm$.000} &\textbf{0.999$\pm$.000}\\
    \hline
    DBSR-cplx46K & Acc  & 0.955$\pm$.012 &   {0.986$\pm$.001} & {0.990$\pm$.001} & {0.990$\pm$.001} & {0.983$\pm$.001} & {0.990$\pm$.002} &0.990$\pm$.001 &\textbf{0.992$\pm$.001}\\
    & Prec   & 0.956$\pm$.010 &   {0.986$\pm$.001} & {0.990$\pm$.001} & {0.987$\pm$.001} & {0.983$\pm$.001} & {0.990$\pm$.002} &0.990$\pm$.001 &\textbf{0.992$\pm$.001}\\
    & F1   & 0.955$\pm$.012 &   {0.986$\pm$.001} & {0.990$\pm$.001} & {0.990$\pm$.001} & {0.983$\pm$.001} & {0.990$\pm$.001} &0.990$\pm$.001 &\textbf{0.992$\pm$.001}\\
    & AUC  & 0.995$\pm$.001 &   {0.997$\pm$.000} & {0.997$\pm$.000} & {0.997$\pm$.000} & {0.997$\pm$.000} & {0.997$\pm$.000} &\textbf{0.998$\pm$.000} &\textbf{0.998$\pm$.000}\\
    \bottomrule
  \end{tabular}
  \label{table:performance}
\end{table*}

\subsection{Effectiveness Analysis}
Table \ref{table:performance} presents a performance comparison between the proposed method and other competing models across five datasets. We utilize a range of metrics to quantify the performance including Accuracy (Acc), weighted precision (Prec), weighted F1 score (F1), and the weighted ROC AUC score (AUC). The results reflect averages from three independent trials for each configuration, with standard deviations noted alongside the $\pm$ symbol. The highest-scoring result for each dataset is denoted in boldface. 

PolygonGNN distinguishes itself with the highest scores in accuracy, precision, and F1 across all datasets, notably improving the F1 score by 18.3\% over the average of other methods in the MNIST-P-2 dataset. This enhancement evidences the method's superior capability, especially when other models show sporadic comparable performance across varied datasets.
Particularly in the MNIST-P-2 dataset, PolygonGNN's proficiency is unmatched, achieving the highest scores in accuracy, precision, F1, and AUC, significantly surpassing competing methods. In the Building-2-C dataset, despite the seemingly modest absolute performance values, PolygonGNN's achievements are substantial, especially given the difficulty competing methods face in making accurate predictions. This challenge arises from the random combination and scaling of buildings within a unified grid, introducing variability in orientation and inter-polygonal relationships even among samples with identical labels. Such complexity underscores the critical need for a model like PolygonGNN, capable of discerning the nuanced inter and inner relationships within a multipolygon.
The Building-2-R dataset presents a scenario akin to Building-2-C but additionally considers the relative positioning of buildings for labeling, which slightly improves the performance of all models.  
PolygonGNN's adaptability is further demonstrated in the Building-S dataset, which focuses on single polygons. Here, PolygonGNN's top performance affirms its robustness and flexibility, effectively handling both single and multipolygon representations through a cohesive data structure. 
DBSR-cplx46K is recognized as a complex and extensive dataset. While various methods demonstrate commendable performance on this dataset, PolygonGNN distinctly stands out by achieving the highest scores across multiple metrics.

\subsection{Ablation Study}
In this section, we explore the advantages of incorporating our sampling method and the Multipolygon-GNN within our framework. To assess the impact of the augmentation introduced by the sampling strategy, we conduct experiments without the sampling technique, labeling this variant as "PolygonGNN w/o S." For the GNN component, we substitute our proprietary GNN with HAN and HGT. The results, presented in Table \ref{table:performance}, indicate that the sampling method substantially enhances performance across nearly all datasets by enriching the training data. Additionally, our custom Multipolygon-GNN outperforms the alternative HAN and HGT, underscoring the importance of developing a specialized GNN tailored for multipolygon encoding.

\subsection{Efficiency Analysis}
\begin{figure}
    \centering
    \includegraphics[width=1\linewidth]{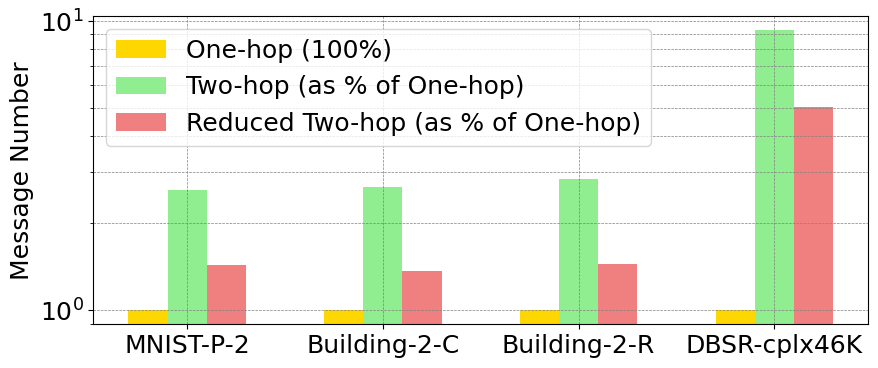}
    \caption{Message number comparison}
    \label{fig:effi}
\end{figure}

To assess the efficiency of our proposed sampling method relative to traditional one-hop and two-hop-based message-passing neural networks, we conducted a comparative analysis of the total number of messages processed by each method. This comparison was performed using the datasets employed in our experiments, excluding the Building-S dataset, which exhibits uniform message counts across all methods due to its singular closed shape structure. The comparative data is presented in Table \ref{tab:message_numbers}.
Furthermore, to quantify the efficiency gains, we computed the ratio of messages processed by the two-hop and our sampling method to those processed by the one-hop method. These ratios, illustrating the relative change in message processing load, are depicted in Figure \ref{fig:effi}. Our findings reveal that the conventional two-hop message-passing approach can result in a message count up to 9.3 times higher than the one-hop method. In contrast, our sampling method significantly reduces this overhead, achieving an average ratio of reduced two-hop to one-hop messages across the datasets at approximately 2.33, thereby demonstrating a substantial improvement in efficiency.

\begin{table}[h]
\small
\centering
\begin{tabular}{cccc}
\hline
Dataset & One-hop & Two-hop & Reduced Two-hop \\ \hline
MNIST-P-2     & 4482           & 11666           & 6410            \\ \hline
Building-2-C     & 2814           & 7520            & 3848            \\ \hline
Building-2-R     & 2743           & 7799            & 3959            \\ \hline
DBSR-cplx46K     & 4712           & 43796            &    23856       \\ \hline
\end{tabular}
\caption{Numbers of calculated messages }
\label{tab:message_numbers}
\end{table}

\begin{figure}
    \centering
    \includegraphics[width=\linewidth]{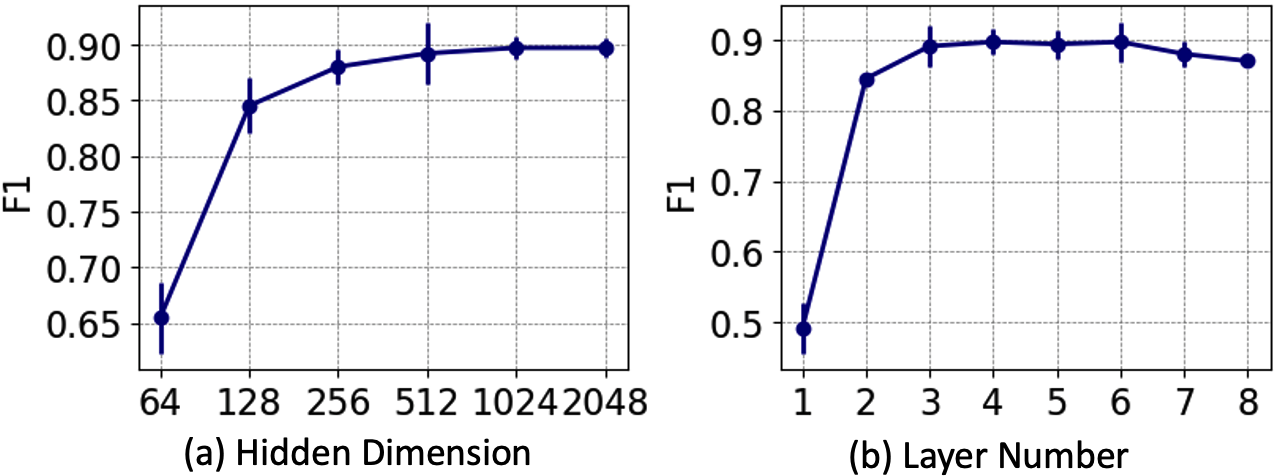}
    \caption{Hyperparameter sensitivity}
    \label{fig:sensitivity}
\end{figure}
\subsection{Hyperparameter Sensitivity}
This section delves into the sensitivity analysis of two critical hyperparameters within our proposed framework: the hidden dimension and the number of graph neural network layers, utilizing the MNIST-P-2 dataset for evaluation.
The impact of the hidden dimension on model performance is illustrated in Figure \ref{fig:sensitivity} (a). Generally, the model exhibits low sensitivity to the hidden dimension size once it surpasses a certain threshold (in this case, 512 for the MNIST-P-2 dataset). Nonetheless, dimensions that are too small may constrict the model's expressive capacity, resulting in suboptimal performance. These findings are consistent with the principles outlined in our Theorem \ref{thm-haus}.
Regarding the number of GNN layers, Figure \ref{fig:sensitivity} (b) reveals a concave curve, indicating a point of diminishing returns. An optimal performance is achieved with approximately 4 GNN layers, suggesting a balance between model depth and efficiency.

\subsection{Embedding Visualization}
\begin{figure*}[htb]
  \centering
  \includegraphics[width=\textwidth]{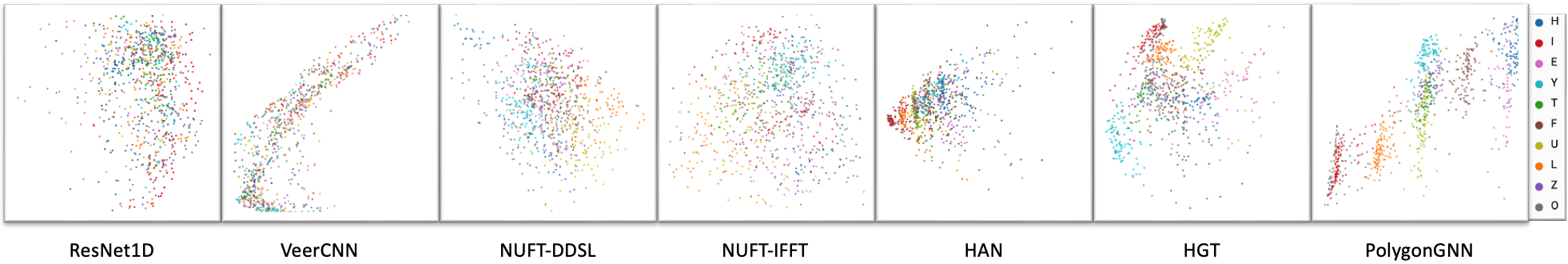}
  \caption{Visualization of learned embeddings in Building-S dataset after PCA dimensionality reduction.}
  \label{fig:embedding}
\end{figure*}
\begin{figure*}[t]
  \centering
  \includegraphics[width=\textwidth]{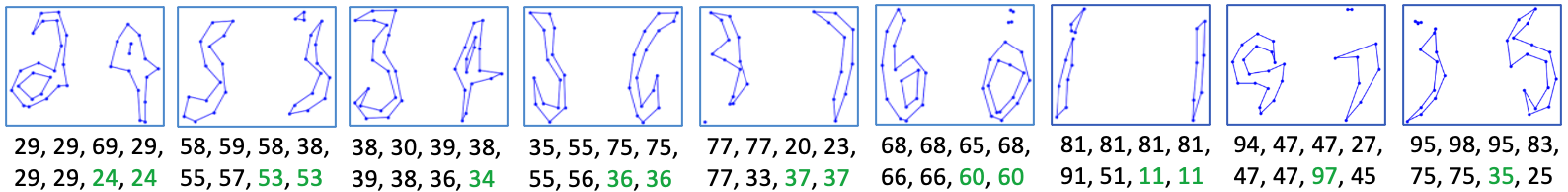}
  \caption{Cases in the test set of the MNIST-P-2 dataset. The predictions made by ResNet1D, VeerCNN, NUFT-DDSL, NUFT-IFFT, HAN, HGT, PolygonGNN w/o S, and PolygonGNN, are displayed beneath each image. Predictions that match the label are highlighted in green.}
  \label{fig:case}
\end{figure*}
We present the visualization of the learned embeddings for multipolygons by different methods in the test set of the Building-S dataset. Each point corresponds to the embedding of an input multipolygon after PCA dimensionality reduction. As depicted in Figure \ref{fig:embedding}, the embeddings generated by our proposed model, including those from its ablation variants, form distinct clusters corresponding to the input categories. In contrast, the embeddings produced by traditional methods appear intermixed. This demonstrates the efficacy of our proposed framework in differentiating between various input multipolygons.

\subsection{Heterogeneous Interaction Weights}
\begin{figure}
    \centering
    \includegraphics[width=\linewidth]{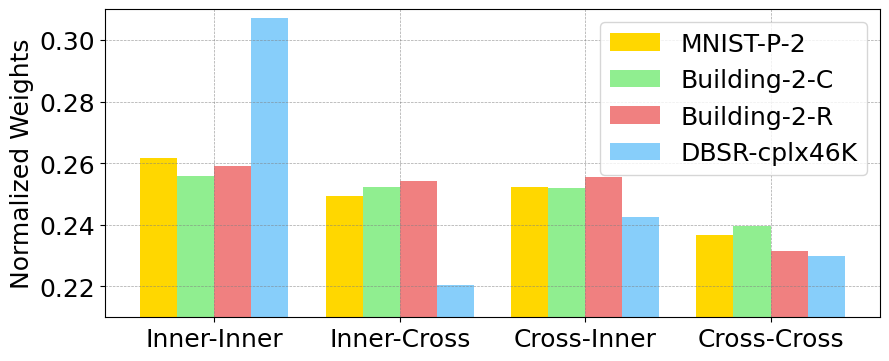}
    \caption{Visualization of interaction weights}
    \label{fig:weight}
\end{figure}

We present the relative weights of the four types of interactions within multipolygons across four datasets in Figure \ref{fig:weight}. Generally, the significance of each interaction varies in accordance with the unique characteristics of each dataset. Specifically, in the DBSR-cplx46K dataset, the weight attributed to the inner-inner interaction is notably higher, whereas the inner-cross interaction is assigned a lower weight. Furthermore, a consistent pattern observed across the datasets is that the cross-cross interaction typically receives the lowest weight, whereas the inner-inner interaction tends to have the highest weight. This is reasonable, given that in the tasks, accurately recognizing the shape of each individual polygon is crucial, and understanding the relationships between polygons is comparatively easier. 
\subsection{Visualization of Prediction Cases}
We delve into the predictive performance of PolygonGNN in the MNIST-P-2 dataset by showcasing several cases in the test set.
In Figure \ref{fig:case}, the predictions from all the methods are listed.
We notice PolygonGNN adeptly handles minor imperfections and downsampling artifacts. 
The first type of noise consists of irregular distortions that result in non-standard shapes. For example, in case "24", the hole in the digit "4" is reduced to a line. The second type of noise involves broken parts, as observed in cases "53", "37", "60", and "11", where a digit is fragmented into several disjointed pieces. This fragmentation confounds comparison methods due to their limited capacity to interpret the relationships among multiple polygons, rendering them unable to identify the digit once it is segmented. In the last two cases, the inputs exhibit extensive damage and are challenging to categorize. However, the unsampled version of our method PolygonGNN w/o S still makes correct predictions. This indicates that adding more visibility edges between multiple polygons may enhance the robustness of the model. Future work could focus on exploring the balance between efficiency and robustness

\section{Conclusion}
This work advances representation learning for polygonal geometries, with a particular focus on multipolygons, through the introduction of a novel framework named PolygonGNN. Central to this framework is the heterogeneous visibility graph, a unified data structure for polygonal geometries, coupled with the development of a heterogeneous geometry representation for these graphs and a custom-designed graph neural network, Multipolygon-GNN. To ensure processing efficiency without compromising the integrity of multipolygon information, a unique heterogeneous spanning tree sampling method is introduced.
The effectiveness of PolygonGNN has been rigorously validated through extensive experiments on both synthetic and real-world datasets, showcasing its robust performance across a variety of scenarios. The representations learned by PolygonGNN are highly discriminative, making them valuable for a wide range of applications.

\begin{acks}
This work was supported by the NSF Grant No. 2432418, No. 2414115, No. 2007716, No. 2007976, No. 1942594, No. 1907805, No. 2318831, Cisco Faculty Research Award, Amazon Research Award.
\end{acks}

\bibliographystyle{ACM-Reference-Format}
\bibliography{software}

\appendix
\section{Proof for Theorem \ref{thm1}}\label{appendix:thm1}
\begin{proof}
The nodes in graph $G$ have a one-to-one correspondence with the vertices of the multipolygon $q$. To reconstruct $q$ from $G$, we need to group the nodes of the graph into their corresponding polygons. For each polygon, we then identify its boundaries and holes. Since The inner edges are directed, we can reconstruct the boundaries and holes by initiating traversal from any node and following the inner edges until the starting node is reached. This traversal allows us to group the nodes into either boundaries or holes of the polygons. Furthermore, nodes along the boundaries of polygons are arranged in a counterclockwise direction, whereas nodes in holes follow a clockwise arrangement. This distinction allows for straightforward discrimination between boundaries and holes through basic geometric computations. Each identified boundary corresponds to a distinct polygon, and the holes identified within these boundaries are allocated to their respective polygons. Consequently, every polygon is reconstituted with its correct shape and internal structure.
Hence, from graph $G$, we can uniquely reconstruct the original multipolygon $q$, ensuring that no information about the polygon's structure is lost.
\end{proof}
\section{Proof for Corollary \ref{thm-aug}} \label{appendix:thm-aug}
\begin{proof}
In the transformation from a multipolygon to a graph, each vertex and edge of the polygon is uniquely mapped to a node and edge in the graph, respectively. The addition of visibility edges does not alter the inherent structure of the original polygon but rather adds supplementary connections between nodes. 
The reconstruction process involves tracing the inner edges to reform the exterior boundaries and holes of the multipolygon, as detailed in the proof of Theorem \ref{thm1}.
Now, consider the set of graphs  $\{G(V, E_{inner}\cup E'_{cross,\omega}, \phi)|\omega \in \Omega \}$, where each $G'$ is a variant of $G$ with a unique subset  $E'_{cross,\omega}$ of visibility edges. Despite the variation in  $E'_{cross,\omega}$  across different  $G'$, the core structure of  $q$ is preserved in  $E_{inner}$. Therefore, regardless of which  $G'$ is considered, the original multipolygon can be uniquely reconstructed. 
\end{proof}
\section{Proof for Lemma \ref{lem}} \label{appendix:lem}
\begin{proof}
We establish a local Cartesian coordinate system with node $v_j$ as the origin and the ray $\overrightarrow{ v_iv_j}$ as the positive x-axis. Define the angle $\theta_{i,j,k}$ as the clockwise rotation from the ray $\overrightarrow{v_jv_i}$ to the ray $\overrightarrow{v_jv_k}$. A positive value of $\theta_{i,j,k}$  indicates a clockwise rotation, while a negative value indicates a counterclockwise rotation. Given this setup, the coordinates of node  $v_k$  relative to  $v_j$  can be calculated using trigonometric relations:
\begin{align*}
    \left\{
    \begin{array}{l}
        x_k = -d_{j,k} \cos(\theta_{i,j,k}), \\
        y_k =  d_{j,k} \sin(\theta_{i,j,k}) 
    \end{array}
    \right.
\end{align*}    
Therefore, by applying these trigonometric relations, we can uniquely determine the coordinates of $v_k$ in the local coordinate system. To obtain the global coordinates, it is essential to first determine the angle of rotation from the global coordinate system to the local one. Once this angle is calculated, we can proceed to apply the necessary rotation and translation operations. These procedures are standard and straightforward in the field of coordinate transformations. To maintain the focus on the central aspect of the lemma, the intricate details of these operations are not elaborated here. The edge types can be directly determined using the type information provided in the tuple. Hence, the lemma is thereby proven.
\end{proof}
\section{Proof for Theorem \ref{lem-thm}}\label{appendix:lem-thm}
\begin{proof}
An equivalent heterogeneous visibility graph is one that represents the same multipolygon, under any translation or rotation transformations. Without loss of generality, we fix the coordinates of two nodes as delineated in Lemma \ref{lem}, then the Cartesian coordinates of a node $v_k$ in a two-hop path $\pi_{i,j,k}$ and the edge types can be determined. Given that the heterogeneous visibility graph $G'$ converted from multipolygon $q$ exhibits strong connectivity, we can iteratively apply this process to determine the coordinates of all nodes in the graph. Starting from any two-hop path,  we can progressively solve for the coordinates of adjacent nodes that are directly connected to the already solved nodes. By sequentially solving for the coordinates of each connected node and expanding the set of nodes with determined coordinates, the Cartesian coordinates of the entire graph can be established.

Accordingly, leveraging the five-tuple heterogeneous geometric representation along with the robust connectivity inherent in the heterogeneous visibility graph, it becomes feasible to ascertain both the coordinates of all nodes in  $G'$ and their connectivity details. This process can be initiated from any selected two-hop path within the graph, provided that the initial two nodes have fixed coordinates. These starting coordinates can be arbitrarily assigned while adhering to the constraints of the five-tuple heterogeneous geometric representation. It's noteworthy that different initializations, which might lead to varying orientations or positions of the graph due to rotation or translation transformations, still correspond to the same multipolygon. Consequently, despite these transformations, the reconstructed graph retains its equivalence to the original heterogeneous visibility graph.  Hence, the theorem is proven.
\end{proof}
\section{Proof for Theorem \ref{thm-haus}}\label{appendix:thm-haus}
\begin{proof}
Since $f: \mathcal{S} \rightarrow \mathbb{R}$ is a continuous set function with respect to Hausdorff distance, $\forall \epsilon_1>0, \exists \delta_1 >0$ such that for any $S,S' \in \mathcal{S}$ with $d_H(S,S')<\delta_1$, we have $|f(S)-f(S')|<\epsilon_1$.
Assume, without loss of generality, that $S$ is a one-dimensional finite set contained within an interval$[a,b]$.
Denote this interval as $\Xi=[a,b]$, we can divide $\Xi$ into $K=\lceil \frac{b-a}{\delta}\rceil+1$ equal subintervals $[a+(k-1)\Delta,a+k\Delta],k=1,2,\dots,K$, where $\Delta=\frac{b-a}{K}$. 
Define a function $r: \mathbb{R} \rightarrow \mathbb{R}$ as $r(x)=a+\lfloor \frac{x-a}{\Delta}\rfloor \Delta$, which maps each $x \in S$ to the lower bound of its respective subinterval. Let $S'=\{r(x):x\in S\}$. By this construction, $d_H(S,S')\leq\frac{b-a}{K}<\delta_1$, hence $|f(S)-f(S')|<\epsilon_1$.

Next, define $\sigma_k: \mathbb{R} \rightarrow [0, +\infty)$ as the Hausdorff distance from any point $x$ to the complement of the $k$-th subinterval in $\Xi$. Specifically, $\sigma_k(x)=d_H(x,\Xi \backslash [a+(k-1)\Delta, a+k\Delta])$. 
Let symmetric function $v_k(S)=\sum_{x \in S} \sigma_k(x)$, indicating whether points of $S$ fall within the $k$-th subinterval.

With these definitions, we construct a mapping function $\tau: [0,+\infty)^K \rightarrow \mathcal{S}$ as $\tau(\mathbf{v})=\{a+(k-1)\Delta: v_k>0\}$, which maps the vector $\mathbf{v}=[v_1,\dots,v_K]$ to a set consisting of the lower bounds of the subintervals occupied by $S$, which exactly equals the set $S'$ constructed above, i.e., $\tau(\mathbf{v}(S))=S'$.

Let $\zeta: \mathbb{R}^K \rightarrow \mathbb{R}$ be a continuous function so that $\zeta(\mathbf{v})=f(\tau(\mathbf{v}))$. Denote $\boldsymbol{\sigma}=[\sigma_1,\dots,\sigma_K]$. Then we have
\begin{align*}
    &|f(S)-\zeta(\sum\{\boldsymbol{\sigma}(x):x \in S\})|\\
    =&|f(S)-f(\tau(\sum\{\boldsymbol{\sigma}(x):x \in S\}))|\\
    =&|f(S)-f(\tau(\mathbf{v}(S)))|\\
    =&|f(S)-f(S')|<\epsilon_1
\end{align*}
The continuous function $\boldsymbol{\sigma}$ can be approximated by a multilayer perceptron, according to the universal approximation theorem. Therefore, We have $|f(S)-\zeta(\sum\{m(x):x \in S\})|<\epsilon$, where $m$ is the MLP function. Considering the method described in Section \ref{sec:hgnn}, we can set $L=1$, making our proposed function $f'$ a sum of the messages from an MLP function. The sum operator is a special case of our method when $L=1$ and the message function is the MLP used above. Thus, we arrive at the conclusion that $|f(S)-\zeta(f'(S))|<\epsilon$. Hence, the theorem is proven.
\end{proof}

\section{Implementation details}\label{appendix:implementation}
Each dataset is randomly split into 60\%, 20\%, and 20\% for training, validation, and testing respectively.

We use CrossEntropyLoss as the loss function for all classification tasks. Adam optimizer and ReduceLROnPlateau scheduler are used to optimize the model. The learning rate is set to 0.001 across all tasks and models. The training batch is set to 64 and the test batch is 128 for datasets except for the DBSR-cplx46K dataset. The training and test batch are 8 for the DBSR-cplx46K dataset. The downstream task model is a four-layer MLP function with batchnorm enabled across all tasks. All models are trained for a maximum of 500 epochs using an early stop scheme.

For the comparison method ResNet1D, VeerCNN, NUFT-DDSL, and NUFT-IFFT, we follow the original settings provided by the authors. For HAN and HGT, we use the implementation of PyG.

For the message encoding function $\Psi$, we use a four-layer MLP function with batchnorm enabled across all tasks. For the geo-embedding function $\varphi$, we leverage a one-layer MLP function with batchnorm enabled across all tasks. 
The hyperparameters we tuned include hidden dimension in {128,256,512,1024}, and the number of GNN layers in {1,2,3,4,8}. We found the best hyperparameters for different datasets are: MNIST-P-2: [1024,4];  Building-2-C: [512,4]; Building-2-R: [512,4]; Building-S: [1024,4]; DBSR-cplx46K: [512,2].

\end{document}